\documentclass[11pt,a4paper]{article}
\usepackage[hyperref]{acl2021}
\usepackage{times}
\usepackage{latexsym}

\usepackage{graphicx}
\usepackage{colortbl}
\usepackage{amssymb}
\definecolor{mygray}{gray}{.9}
\usepackage{scalerel,xparse}
\NewDocumentCommand\emojismile{}{
    \scalerel*{
        \includegraphics{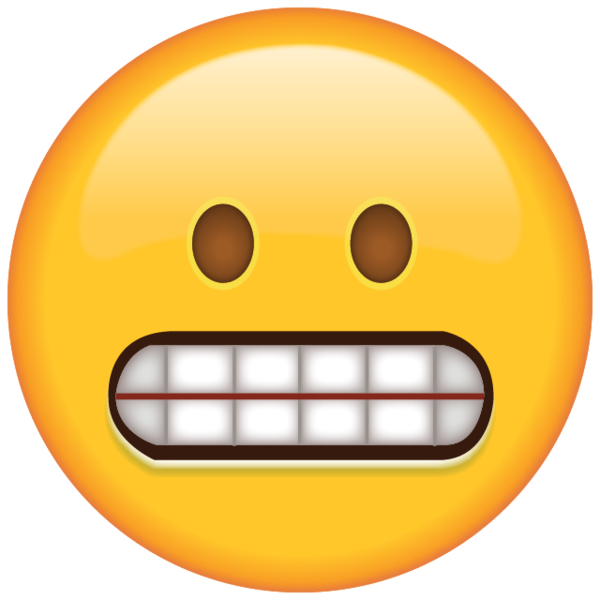}
    }{X}
}

\usepackage{microtype}

\aclfinalcopy 

\title{\textsc{DialogSum}: A Real-Life Scenario Dialogue Summarization Dataset}

\author{
 Yulong Chen$^{\spadesuit \heartsuit}$\hspace{0.5mm}, 
 Yang Liu$^\clubsuit$\hspace{0.5mm}, 
 Liang Chen$^{\blacklozenge}$\hspace{0.5mm}, 
 Yue Zhang$^{\heartsuit \diamondsuit}$\hspace{0.2mm}\hspace{1.5mm} \\
 $^\spadesuit$ Zhejiang University\\
 $^\heartsuit$ School of Engineering, Westlake University\\
 $^\clubsuit$ Institute for Language, Cognition and Computation, University of Edinburgh\\
 $^\blacklozenge$ College of Software, Jilin University\\
 $^\diamondsuit$ Institute of Advanced Technology, Westlake Institute for Advanced Study\\ 
 \textit{yulongchen1010@gmail.com} \quad\textit{inf.yangl@outlook.com}\\
 \textit{chenliang5518@mails.jlu.edu.cn} \quad\textit{yue.zhang@wias.org.cn} \\
}

\date{}

\begin{document}
\maketitle
\begin{abstract}
Proposal of large-scale datasets has facilitated research on deep neural models for news summarization. 
Deep learning can also be potentially useful for spoken dialogue summarization, which can benefit a range of real-life scenarios including customer service management and medication tracking. 
To this end, we propose \textsc{DialogSum}, a large-scale labeled dialogue summarization dataset. 
We conduct empirical analysis on \textsc{DialogSum} using state-of-the-art neural summarizers. 
Experimental results show unique challenges in dialogue summarization, such as spoken terms, special discourse structures, coreferences and ellipsis, pragmatics and social common sense, which require specific representation learning technologies to better deal with.
\end{abstract}

\section{Introduction}
Text summarization is the task of automatically generating a concise, salient, coherent and fluent summary of a given set of documents~\cite{radev2002introduction}.
Thanks to the advance in neural network models and the availability of large-scale labeled datasets, recent research has achieved promising progress on summarizing monologic texts such as news articles~\cite{paulus2017deep, gehrmann2018bottom, liu2019text, liu2020noisy}, patents~\cite{subramanian2019extractive} and academic papers~\cite{koncel2019text}.
\begin{figure}[t!]
    \centering
    \includegraphics[width=1\columnwidth]{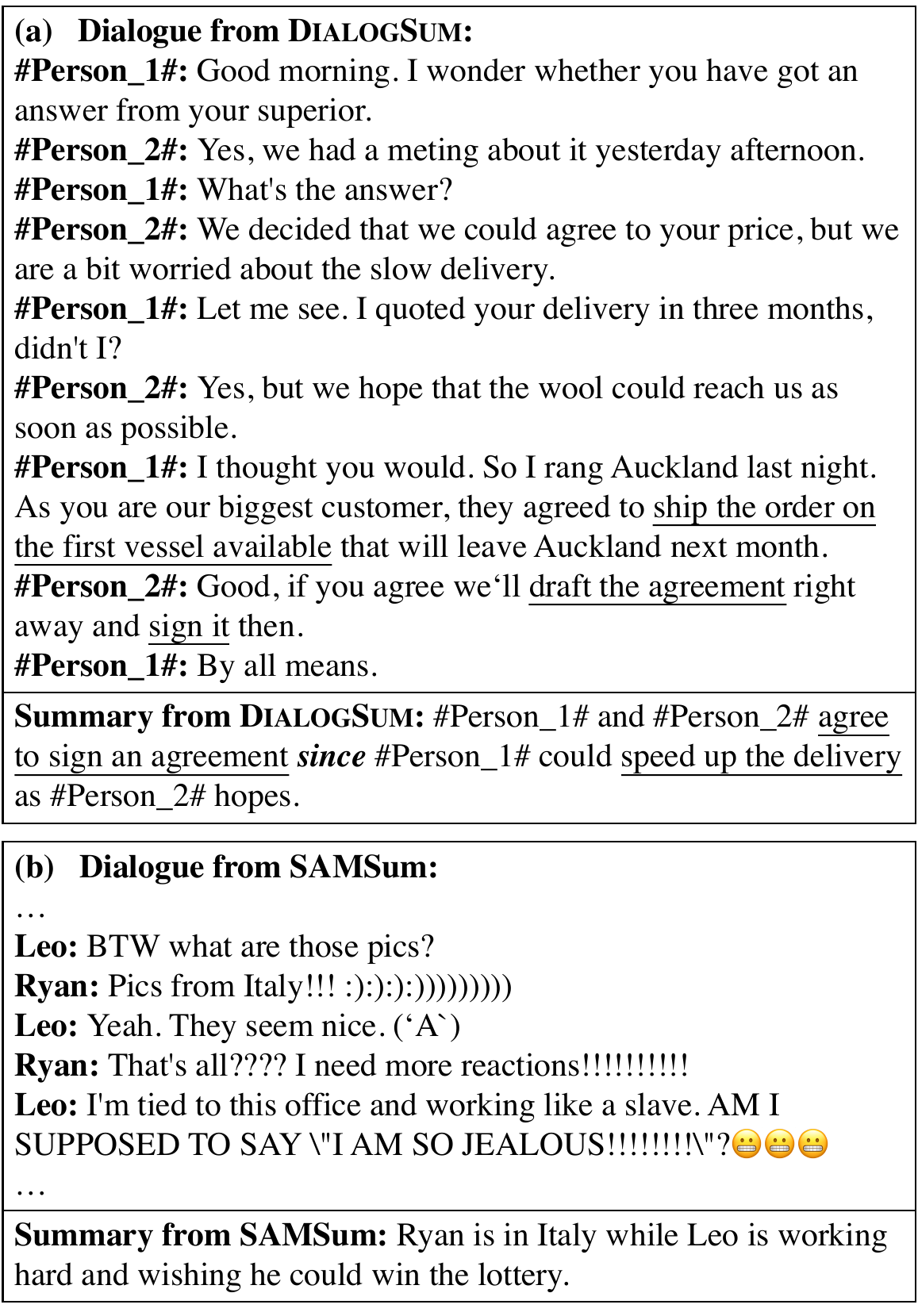}
    \caption{An example from \textsc{DialogSum} dataset compared with an example from SAMSum dataset.}
    \label{example}
\end{figure}

\begin{table*}[]
\setlength\tabcolsep{3.5pt}
\small
\centering
{
\begin{tabular}{l|cccccccc}
\hline
\textbf{Datasets} & \textbf{Lan. style} & \textbf{Domain} & \textbf{Scenario} & \textbf{Dialogs}  & \textbf{Data size} &  \textbf{\#Tokens/dial.} & \textbf{\#Tokens/turn} & \textbf{\#Comp. rate}\\
\hline
AMI & spoken & single & meeting& 137 &100hrs (video) & 4,757 & 16.5 & 0.07 \\
SAMSum  & written & multiple & online & 16,369 & 1.5M (token) & 94 & 8.4  & 0.30\\
\textsc{DialogSum} & spoken & multiple & daily life & 13,460 & 1.8M (token) & 131 & 13.8 & 0.18\\
\hline
\end{tabular}
}
\caption{Comparison between \textsc{DialogSum} and other public dialogue summarization datasets.
Lan. stands for language.
Dial. stands for dialogue.
\# stands for the average result.
Comp. stands for compression.
The compression rate is the ratio of the length of the summary divided by the length of the original text.
}
\label{comparision}
\end{table*}

However, dialogue, as an important channel for achieving communicative intents~\cite{bender2020climbing}, has received significantly less attention from the summarization research community.
One main reason is the paucity of a suitable summarization dataset built on dialogue texts.
Most existing research uses the AMI meeting corpus~\cite{carletta2005ami}, which consists of $137$ dialogues obtained from virtual multi-party meeting recordings. 
However, research on the corpus is limited to its small scale.
SAMSum~\cite{gliwa2019samsum} is a recently released \textit{written} online dialogue summarization dataset, which contains $16$k online chats with corresponding summaries. However, it focuses on conversations via messenger apps, which are rather short (around $94$ tokens per conversation) and their language style and topics also differ from \textit{spoken daily dialogues}.

A comparison between the real-life scenario dialogue and online chat is shown in Figure~\ref{example}. 
Online-chat messages contain unique tokens (e.g., ``BTW''), emoticons (e.g., ``:)'') and emojis (e.g., ``\emojismile''). 
In contrast, daily conversations have a different and more formal style. 
In addition, real-life dialogues have more diverse task-oriented scenarios and topics compared to online chit-chats.
For example, online-chat messages in SAMSum are about leisure and social chats, but real-life dialogues contain business negotiation (Figure~\ref{example}(a)).
Intuitively, automatically summarizing such dialogues can help a business find common needs or complaints from customers. 
With the rise of personal assisting chatbots, summarizing dialogues from different aspects of daily life can also be useful for personal record management and other applications.

We introduce Real-Life Scenario Dialogue Summarization (\textsc{DialogSum}), a large-scale summarization dataset for dialogues.
Dialogue data for \textsc{DialogSum} are collected from three public dialogue corpora, namely Dailydialog~\cite{li2017dailydialog}, DREAM~\cite{sun2019dream} and MuTual~\cite{cui2020mutual}, as well as an English speaking practice website. 
These datasets contain face-to-face spoken dialogues that cover a wide range of daily-life topics, including schooling, work, medication, shopping, leisure, travel.
Most conversations take place between friends, colleagues, and between service providers and customers.
We clean and preprocess the dialogue data into a unified format, and ask annotators to summarize them from an observer perspective.
Topics are also manually labeled for each dialogue. 
An example of \textsc{DialogSum} is shown in Figure~\ref{example}(a), where the summary expresses the main content in a business conversation.

The contribution of \textsc{DialogSum} can be stated from two perspectives. 
First, from the perspective of downstream applications, summarizing daily spoken dialogues can be useful for both business and personal uses. 
Dialogue summaries can also be useful for personal assistants to keep track of important events as such business negotiation. 
Second, from the method perspective, \textsc{DialogSum} has a larger scale of long dialogue data, which can facilitate the study of dialogue summarization using neural network models.
The number of dialogues in \textsc{DialogSum} is orders of magnitude larger than in AMI, which can be useful for training large neural network models for dialogue summarization.
The average length of dialogues in \textsc{DialogSum} is $39.8\%$ longer than in SAMSum.
To our knowledge, we are the first to release a large-scale real-life scenario dialogue summarization dataset.

We empirically investigate the performance of state-of-the-art neural summarization models on \textsc{DialogSum}, comparing the characteristics of the spoken daily dialogue summarization dataset with standard news summarization benchmarks and the online chat summarization benchmark SAMSum.
Experimental results show that \textsc{DialogSum} is more amenable to abstractive summarizers, while being relatively more challenging compared to the existing summarization datasets. 
We find that main difficulties arise from discourse structures in multi-turn dialogues, as well as the need for book-keeping both entities and events mentioned in turns of utterances. 
We release our dataset at \url{https://github.com/cylnlp/DialogSum}.

\section{The \textsc{DialogSum} Dataset}

\subsection{Dialogue Data Preparation}
\paragraph{Data Collection} 
DailyDialog is a dataset consisting of 13k multi-turn dialogues, obtained from websites that aim to help English learners to practice English speaking. 
DREAM and MuTual are dialogue understanding datasets, consisting of 6k and 9k speech transcripts, respectively, both collected from online English listening exam materials.
In order to further increase the diversity of data, we crawl additional dialogues from another English speaking practice website\footnote{http://www.tingroom.com} which aims to provide English learners with conversation examples in real life practical circumstances, such as business negotiation and banking services. 

Although dialogues of \textsc{DialogSum} are from different sources, they all share important characteristics that are in line with what we expect.
First, as mentioned earlier, these dialogues are under rich real-life scenarios. 
Unlike chitchats, these conversations have clear communication patterns and intents, making them more suitable and valuable to serve as summarization sources~\cite{carletta2005ami}.
Moreover, their multi-turn dialogue lengths are within a reasonable scale and are longer than chitchats\footnote{The average numbers of tokens for multi-turn dialogues are: Dailydialog: $118.8$, DREAM: $124.6$, MuTual: $136.1$.}, which comforts the purpose of automatic summarization. 
Greater lengths also indicate these dialogues contain more events and discourse relations between them. 
Properly selecting vital events and identifying their relations make summarizing these dialogues more challenging.

\begin{table}[]
\setlength\tabcolsep{5pt}
\small
\centering
{
\begin{tabular}{l|ccc}
\hline
\textbf{Source} & \textbf{Num. dial.} & \textbf{Tokens} & \textbf{\% in \textsc{DialogSum}} \\
\hline
DailyDialog & 7837 & 980,398 & 58.22  \\
DREAM  & 2028 & 301,098  & 16.94\\
MuTual & 1870 & 260,139 & 13.89 \\
Crawled & 1725 & 238,902 & 12.82\\
\hline
\end{tabular}
}
\caption{Proportions of dialogue sources in \textsc{DialogSum}.
}
\label{proportiontable}
\end{table}

\paragraph{Data Cleaning and Pre-Processing} We delete non-English characters, correct typos and grammatical errors, and further filter out duplicated data based on text similarity.
After deduplicating, proportions of the data sources are summarized in Table~\ref{proportiontable}.
Because of different data processing methods and annotation procedures, original dialogues in DailyDialog, DREAM and MuTual are in different formats. 
We follow previous work~\citep{li2017dailydialog,zhang2018personalizing,budzianowski2018multiwoz,dinan2018wizard} and preprocess them into a bi-turn dialogue flow, merging continuous turns of the same speaker into one utterance.
Also, we add tags (e.g. \texttt{\#Person\_1\#} and \texttt{\#Person\_2\#} in Figure~\ref{example}(a)) before each dialogue turn, to distinguish speakers.
The final \textsc{DialogSum} dataset contains 13,460 dialogues, which are divided into training (12,460), validation (500) and test (500) sets.

\subsection{Annotation}\label{annotation}
We ask annotators to write dialogue summaries based on following criteria: the summary should 
(1) convey the most salient information of the dialogue and; (2) be brief (no longer than 20\% of the conversation length) and; (3) preserve important named entities within the conversation and; (4) be written from an observer perspective and; (5) be written in formal language. 

We require our annotators to pay extra attention to the following aspects.

\textbf{Tense Consistency:} Annotators should take the moment that the conversation occurs as the \textit{present} time, and choose a proper tense to describe events \textit{before} and \textit{after} the ongoing conversation.

\textbf{Discourse Relation:} If summarized events hold important discourse relations, particularly causal relation, annotators should preserve the relations if they are also in the summary.

\textbf{Emotion:} Different from newspaper and academic articles, social conversations in \textsc{DialogSum} are often implied with emotions. 
Therefore, we ask annotators to explicitly describe important emotions related to events in the summary.

\textbf{Intent Identification:} Rather than merely summarizing the consequences of dialogues, annotators should also describe speakers' intents in summaries, if they can be clearly identified.

In addition to the above, annotators should use person tags to refer to different speakers if real names cannot be detected from the conversation.
Annotators are also asked to write a short (around 3 tokens) topic for each dialogue.
Appendix~\ref{a_topic} shows the list of topics.




\begin{table}[!t]
\centering
\setlength\tabcolsep{3.5pt}
\small
{
\begin{tabular}{l|ccc}
\hline
\textbf{Human Annotated Summary}& R1 & R2 & RL\\
\hline
Summary1 to Summary2 & 52.90 & 26.01 & 50.42\\
Summary1 to Summary3 & 53.85 & 27.53 & 51.65\\
Summary2 to Summary3 & 53.30 & 26.61 & 50.44\\
\hline
Average & 53.35 & 26.72 & 50.84\\
\hline
\end{tabular}
}
\caption{ROUGE scores between three human annotated summaries in test set.}
\label{between-rouge}
\end{table}

\begin{table*}[!t]
\centering

\setlength\tabcolsep{3.5pt}
\small
{
\begin{tabular}{l|cccc|ccc|ccc|ccc}
\hline
{\textbf{Dataset}}& \multicolumn{4}{c|}{\textbf{\% of novel $n$-grams}} & \multicolumn{3}{c|}{\textbf{\textsc{lead}}} &  \multicolumn{3}{c|}{\textbf{\textsc{longest}}} & \multicolumn{3}{c}{\textbf{\textsc{ext-oracle}}}\\  
& unigram & bigram &trigram& 4-gram & R1 & R2 & RL & R1 & R2 & RL & R1 & R2 & RL\\
\hline
CNN & 16.75 & 54.33 & 72.42 & 80.37 & 29.15 & 11.13 & 25.95 & - & - & - & 50.38 & 28.55 & 46.58\\
DailyMail & 17.03 & 53.78 & 72.14 & 80.28 & 40.68 & 18.36 & 37.25 & - & - & - & 55.12 & 30.55 & 51.24\\
NY Times  & 22.64 & 55.59 & 71.93 & 80.16 & 31.85 & 15.86 & 23.75 & - & - & - & 52.08 & 31.5 & 46.72\\
XSum      & \textbf{35.76} &\textbf{ 83.45} & \textbf{95.50} & \textbf{98.49} & \textbf{16.30} & \textbf{1.61}  & \textbf{11.95} & - & - & - & \textbf{29.79} & \textbf{8.81} & \textbf{22.65}\\
\hline
SAMSum  & 32.63 & 77.22 & 89.27 & 94.83 & 31.41 & 8.70  & 30.41 & 32.13 & 10.13 & 29.11 & 44.60 & 17.37 & 39.38\\
\rowcolor{mygray} \textsc{DialogSum}     & 26.28 & 76.94 & 89.16 & 94.53 & 27.52 & 6.78  & 27.31 & 24.15 & 6.25  & 22.73 & 37.90 & 13.88 & 34.04\\
\hline
\end{tabular}
}
\caption{Corpora statistics and extractive methods on CNN/DailyMail, NY Times, XSum, SAMSum and \textsc{DialogSum}. Part of results is from \citet{narayan2018don}. 
All results are computed on test sets. 
For \textsc{DialogSum}, the results are the average of multi-reference results.}
\label{bias}
\end{table*}

\subsection{Quality Control}

To ensure quality, before formal annotation, we ask annotators to annotate training samples until they pass our examination and meet our requirements.
After annotation, we check summaries by cross-validation between different annotators twice.
During the checking process, bonus is paid to checkers who find unqualified summaries, and penalty is given to annotators whose annotation is found with mistakes. 
In case of appeal, we make the final decision.
After the second checking, we sample $10\%$ summaries and manually check the samples ourselves. 
If errors are found in an annotation batch, we ask corresponding annotators to self-check and re-annotate the whole batch and repeat this checking and sampling processes. 

To further control the quality, and to analyze inter-annotator agreement, for each dialogue in the test set, we provide three summaries written and checked by different annotators. 
For each test dialogue, we compare its three summaries and compute their pair-wise \textsc{ROUGE}~\citep{lin2004rouge} scores.
Table~\ref{between-rouge} reports their averaged $F_1$ scores of \textsc{ROUGE-1} (R1), \textsc{ROUGE-2} (R2) and \textsc{ROUGE-L} (RL).
We see \textsc{R2} is relatively low while \textsc{RL} is high, which suggests that annotators' usage of language is variable, but the main content and logical order are mostly the same.


\subsection{Characteristics of \textsc{DialogSum}}

We empirically compare \textsc{DialogSum} with existing news summarization datasets and SAMSum.
CNN/DailyMail~\citep{hermann2015teaching}, NY Times~\citep{sandhaus2008new} and XSum~\citep{narayan2018don} are large-scale summarization datasets from the news domain, written in a monologic structure.
XSum is a dataset designed specifically for abstractive summarization.

First, we compare the percentages of novel $n$-grams in the reference summary against the source document/dialogue.
This intuitively reflects the level of abstraction of annotated summaries. 
As shown in Table~\ref{bias}, except for XSum, which is designed to be highly abstractive, dialogue-based summarization datasets contain more novel $n$-grams in the summaries.
We also find that the percentage of novel unigrams in \textsc{DialogSum} is $26\%$, $6\%$ lower than in SAMSum, but novel bigrams, trigrams and 4-grams are about the same as SAMSum.
We believe that the relatively lower novel unigram proportion in \textsc{DialogSum} compared to SAMSum is because of our pre-processing and annotation criteria. 
SAMSum's summaries include real names, third-person singular pronouns, which can be diverse across the dialogues. 
In contrast, \textsc{DialogSum} uses tags such as \texttt{\#Person\_1\#} to refer to persons whatever they are subjective, objective, or possessive.
This constrains the proportion of novel unigrams to be lower.

Second, we compare the datasets using several extractive summarization methods. 
Following previous summarization work~\citep{liu2019fine, subramanian2019extractive}, we report \textsc{R1}, \textsc{R2} and \textsc{RL} here.
\textsc{lead} creates summaries by selecting the first $n$ sentences from source texts.
\textsc{longest} is designed for dialogue summarization~\citep{gliwa2019samsum}. 
It selects the $n$ longest utterances as a summary, which gives better ROUGE scores than \textsc{lead} on SAMSum.
\textsc{ext-oracle} creates summaries by choosing $n$ sentences that have the highest \textsc{ROUGE} against reference summaries.
It can be viewed as an upper bound for extractive summarization.
We report results of \textsc{lead-3}, \textsc{longest-3} and \textsc{ext-oracle-2} on SAMSum, and \textsc{lead-2}, \textsc{longest-2} and \textsc{ext-oracle-2} on \textsc{DialogSum}, where $n$ is searched for each dataset in range of $1$ to $6$.



\begin{table*}[!t]
\small
\centering
\begin{tabular}{l|ccc|ccc|ccc|ccc}
\hline
{\textbf{Model}} & \multicolumn{3}{c|}{\textbf{CNNDM}}&\multicolumn{3}{c|}{\textbf{XSum}}& \multicolumn{3}{c|}{\textbf{SAMSum}} & \multicolumn{3}{c}{\textbf{\textsc{DialogSum}}} \\  
& R1 & R2 & RL & R1 & R2 & RL & R1 & R2 & RL & R1 & R2 & RL \\
\hline
Transformer &40.21 &17.76 & 37.09& 29.41 & 9.77 &23.01 & 37.20 & 10.86 & 34.69  & 35.91  & 8.74 & 33.50 \\
\hline

$\textsc{UniLMv2}_\textsc{base}$ & 43.16 & 20.42 & 40.14 & 44.00 & 21.11 & 36.08 & 50.53 & 26.62 & 48.81 & 47.04 & 21.13 & 45.04
\\
$\textsc{BART}_\textsc{large}$  & \textbf{44.16}  & \textbf{21.28}&  \textbf{40.90}  & \textbf{45.14}  & \textbf{22.27} & \textbf{37.25} & \textbf{53.12} & \textbf{27.95} &\textbf{49.15}  & 47.28 & 21.18 &  44.83  \\
\hline
\end{tabular}
\caption{Results of abstractive models on CNNDM, XSum, SAMSum and \textsc{DialogSum}. 
For \textsc{DialogSum}, we give the average of multi-reference results.
}
\label{results}
\end{table*}

The results are shown in Table~\ref{bias}. 
In terms of \textsc{lead}, \textsc{DialogSum} sees the lowest R1 and R2 except for XSum, showing that it is in nature a highly abstractive summarization dataset. 
SAMSum is less abstractive than \textsc{DialogSum} by all \textsc{ROUGE} scores, which is likely because the compression rate of SAMSum ($0.30$) is higher than \textsc{DialogSum} ($0.18$) (Table~\ref{comparision}). 
The higher compression ratio suggests the summary contains denser information in the original text.
The same conclusion can be found by using the \textsc{longest} method. By using the \textsc{ext-oracle} method, we find that \textsc{DialogSum} is the most challenging dataset for extractive summarizers except for XSum, which is carefully designed for evaluating abstractive summarizers.

\section{Experiments}
We experiment with several abstractive summarization baselines to further understand the characteristics and challenges of \textsc{DialogSum}.
Following \citet{gliwa2019samsum}, we concatenate utterances of a dialogue as the input. 
For pretrained models, we only finetune them on corresponding datasets.

\subsection{Models}
\paragraph{Transformer} 
We take Transformer~\citep{vaswani2017attention} as a non-pretrained abstractive baseline. 
For dialogue summarization, we follow~\citet{gliwa2019samsum}, using the same hyper-parameters for news summarization\footnote{https://opennmt.net/OpenNMT-py/examples/Summarization.html}, but changing the minimum length to 15.
We train the 6-layer Transformer model with Adam~\citep{kingma2014adam} for 100,000 steps. 
Copy attention mechanism is applied and the dropout rate is set to 0.1.



\paragraph{\textsc{UniLMv2}}
\textsc{UniLMv2}~\citep{bao2020unilmv2} is a recently released pretrained language model for autoencoding and partially autoregressive language modeling.
Here we use $\textsc{UniLMv2}_\textsc{base}$ as a strong abstractive model.
For dialogue summarization, we train the model with Adam for 100,000 steps with 2,000 warmup steps and learning rate is set to $1.5e^{-5}$. 

\paragraph{BART}
BART~\citep{lewis2019bart} is an encoder-decoder Transformer model pretrained on a large corpus using a denoising autoencoder task. 
We use the large version of BART and finetune it with 5,000 training steps/200 warmup steps for dialogue summarization.
Learning rate is set to $3e^{-5}$.



\subsection{Results}
Table~\ref{results} presents the experimental results. 
In general, we find that non-pretrained abstractive models outperform \textsc{lead} (Table~\ref{bias}), and the best results are achieved by pretrained models, despite the fact that $\textsc{BART}_\textsc{large}$ and $\textsc{UniLMv2}_\textsc{base}$ are pretrained on monologic texts.
\paragraph{Extractive Summary vs Abstractive Summary}
Transformer gives similar results on CNNDM and better results on XSum, SAMSum and \textsc{DialogSum} compared to \textsc{lead}, and pretrained models show better performance than \textsc{ext-oracle} on all datasets except for CNNDM.
In particular, pretrained models outperform Transformer by $13.07\sim 14.24\%$ RL on XSum, $14.12\sim 14.46\%$ RL on SAMSum, and $11.33\sim 11.54\%$ on \textsc{DialogSum}, while only $3.05\sim 3.81\%$ on CNNDM.
We believe that it is because CNNDM is a highly extractive dataset (Section 2.4).
The key to summarizing CNNDM is to correctly understand intersentence relations within long documents, and extract important sentences.
In contrast, XSum, SAMSum and \textsc{DialogSum} are more abstractive, which require a model to paraphrase.
And the strong generation capability of pretrained models can bring great improvements on them.
We also see that, for abstractive datasets, model performance decreases as document length grows (Avg. length: SAMSum - $93.8$, \textsc{DialogSum} - $131.1$, Xsum - $431.1$) and compression rate decreases (Comp. rate: SAMSum - $0.30$, \textsc{DialogSum} - $0.18$, XSum - $0.05$).
This explains why SAMSum is the easiest dataset.

\paragraph{Spoken vs Written}


All three models perform better on dialogue summarization datasets, compared with XSum.
This can be potentially because XSum is naturally highly abstractive, and thus more challenging.
We also compare improvement brought by pretrained models that are trained on large written texts.

Still in Table~\ref{results}, the improvement on \textsc{DialogSum} is the least.
$\textsc{BART}_\textsc{large}$ outperform Transformer by $15.73\%$ R1 on XSum, $15.92\%$ R1 on SAMSum, but $11.37\%$ R1 on \textsc{DialogSum}.
It demonstrates that SAMSum has overall more written style than \textsc{DialogSum}, and also suggests that dialogue and monologue are different.
This can be explained by the design of written app-chat annotation  in SAMSum~\citep{gliwa2019samsum}.

\paragraph{\textsc{DialogSum} vs SAMSum}
As shown in Table~\ref{results}, model performance is steadily lower on \textsc{DialogSum} than SAMSum.
As stated, \textsc{DialogSum} is more abstractive, open-domain, and spoken analogous.
One more possible reason for the lower performance on \textsc{DialogSum} is the longer input size.
To better quantify the difference between these two dialogue summarization datasets, we further evaluate Transformer trained on \textsc{DialogSum} when tested on the SAMSum, and vice versa.
As shown in Table~\ref{Transfer}, the performance of Transformer drops greatly when traiend on \textsc{DialogSum} and tested on SAMSum, and vice versa. 
This shows that the two datasets have substantial differences. 
In addition, Transformer trained on \textsc{DialogSum} performs better than on SAMSum, and shows lower performance drop, suggesting that \textsc{DialogSum} can provide more generalization ability for training dialogue summarization models.

\begin{table}[!t]
\centering
\small
{
\begin{tabular}{l|ccc}
\hline
\textbf{Trans. Test} & R1 & R2 &  RL \\
\hline
S2D & 31.72(-5.48) & 6.25(-4.61) & 29.72(-4.97)\\
\hline
D2S & 31.74(-4.17) & 5.93(-2.81) &29.79(-3.71)\\
\hline
\end{tabular}
}

\caption{Difference between \textsc{DialogSum} (D) and SAMSum (S). Trans. stands for Transferred.
}
\label{Transfer}
\end{table}


\section{Human Evaluation}\label{evaluationsection}
To better understand \textsc{DialogSum}, we take a deeper investigation into the outputs of Transformer and \textsc{UniLMv2} on \textsc{DialogSum} by conducting human evaluation from multiple aspects.


\paragraph{Fluency, Consistency, Relevance and Coherence}
First, following~\citet{kryscinski2019neural,kryscinski-etal-2020-evaluating}, we implement human evaluation from four dimensions.
\textit{Fluency} evaluates the quality of individual generated sentences, \textit{Consistency} evaluates the factual alignment between the source text and generated summary, \textit{Relevance} evaluates the importance of summary content, and \textit{Coherence} evaluates the collective quality of all sentences.

We randomly select $50$ dialogues and their summaries from \textsc{DialogSum} test, and ask a judge to give scores in scale from $1$ to $5$ along the four mentioned dimensions. 
The higher, the better.
The judge also gives scores to human-annotated summaries to evaluate their quality.
\begin{table}[]
\centering
\setlength\tabcolsep{3pt}
\small
{
\begin{tabular}{l|cccc}
\hline
\textbf{Summary} & \textbf{Fluency} & \textbf{Cons.} & \textbf{Relevance} &\textbf{Coherence}  \\
\hline
Summary 1 & 5 & 5 & 4.96 & 5\\
Summary 2 & 5 & 5 & 4.98 & 5\\
Summary 3 & 5 & 5 & 5 & 5\\
Avg. & 5 & 5 & 4.98 & 5\\
\hline
Transformer & 4	 &2.08 & 2.3 & 3.84 \\
$\textsc{UniLMv2}_\textsc{base}$   & 4.8 & 3.84 & 4.06 & 4.34 \\
\hline
\end{tabular}
}
\caption{Human evaluation on human annotated summaries and model generated Summaries. Cons. stands for Consistency. Summary 1 - Summary 3 correspond to three summaries of a dialogue.}
\label{fluent}
\end{table}
As shown in Table~\ref{fluent}, human annotated summaries receive the best scores from all dimensions.
$\textsc{UniLMv2}_\textsc{base}$ has steadily better scores than Transformer, but lower than human.
Model-generated summaries have the highest scores on Fluency, while lowest on Consistency. 
It suggests that although model-generated summaries are grammatical and fluent, they still contain factual errors.

\paragraph{Discourse Relation}
Reasonable summaries should convey important relations between main events, and identifying discourse relations and using proper phrases to express them in summaries can be challenging for summarization systems~\citep{xu2020discourse}.
Take Figure~\ref{example} (a) for example, the human annotated summary connects two main events (underlined) using ``\textit{since}'' to express their causal relation explicitly.
However, the causal relation between those two events is not explicitly expressed in the dialogue, and the distance between them is long. 
Multiple turns usually correspond to more complicated discourse structure and relation.
Also, similar with~\citet{chen2020multi}, we find that model performance decreases when the number of dialogue turns grows (See Appendix~\ref{turnnnnn}).

    

To better evaluate model ability to disambiguate discourse relations in \textsc{DialogSum}, we first collect discourse connectives from Penn Discourse Treebank~\citep{prasad2008penn}, and check whether these connectives are included in summaries in the testset.
If the three reference summaries of a dialogue all contain connectives, we assume that the dialogues have strong discourse signals.
We choose $70$ dialogues from \textsc{DialogSum} in this way.

We then ask linguists who specialize in discourse to evaluate model outputs and give scores from $\{-1, 0, 1\}$, where $1$ means that the generated descriptions of main events are reasonable and contain correct discourse connectives, $0$ means that the descriptions are good but contain no discourse connectives and $-1$ means that the description is either incorrect or contains incorrect connectives.
We ask the linguists to focus only on clauses or phrases that are essential to discourse relations, and ignore syntactic errors.
We report the distribution of annotated scores in Table~\ref{disco}.

\begin{table}[]
\centering

\setlength\tabcolsep{3pt}
\small
{
\begin{tabular}{l|cccc|ccc}
\hline
{\textbf{Model}}  & \multicolumn{4}{c|}{\textbf{Human Scores}} & \multicolumn{3}{c}{\textbf{ROUGE Scores}}  \\
& -1 & 0 & 1 & Avg.& R1 & R2 & RL\\
\hline
Transformer & 80\%& 17\% &3\%&-0.77 & 34.35 & 7.01 & 31.13 \\
\textsc{UniLMv2}  & 43\%& 37\%&20\%& -0.23 & 43.78 & 17.91 & 40.97 \\
\hline
\end{tabular}
}
\caption{Human evaluation on discourse relations, with corresponding \textsc{ROUGE} scores on the sub-test set.
Avg. stands for the averaged score here.
}
\label{disco}
\end{table}

We can see that the most summaries generated by Transformer are scored as $-1$, and their average score is $-0.77$, close to $-1$. 
This means that Transformer is not only incapable of identifying discourse relations but also incapable of generating the main events correctly. 
\textsc{UniLMv2} has a relatively smooth distribution over three categories and a better average score of $-0.23$, which is closer to $0$, suggesting that \textsc{UniLMv2} can mostly choose important events amongst the conversation. 
But the $-1$ still holds most proportion and its average result is still far from $1$, indicating its incapability of understanding relations between events. 

Compared to the full test set, the model performance on this sub-set generally decreases ($1.56 \sim 3.26\%$ lower of R1, $1.73 \sim 3.22\%$ of R2, $2.37 \sim 4.07\%$ of RL), which also suggests complicated discourse relations between events make summarization more difficult.
The results indicate that further research is necessary for better representing dialogue discourse structures in order to obtain more reliable summarization systems.

\paragraph{Coreference Information}
To evaluate model's ability to distinguish different interlocutors, we ask a judge to evaluate whether interlocutors' names and their conversation actions/contents are correctly associated in the 50 randomly selected data, and give scores from $\{-1, 0, 1\}$, where $1$ means that all names and actions/content in the summary are associated correctly, $0$ means partial incorrectly, and $-1$ means all incorrectly.
Here, we only focus on coreference information in generated summaries, and ignore other errors, such as incorrect syntax or failing to summarize salient information.

We report the distribution of annotated scores in Table~\ref{coref}. 
Most Transformer generated summaries are annotated as $-1$ and the average result is close to $-1$, suggesting that Transformer cannot generate clauses that express the same relation between arguments and predicates in original dialogues.
The $\textsc{UniLMv2}_\textsc{base}$ has more $0$-scored summaries, and the result is much higher, yet closer to $0$, which indicates that although $\textsc{UniLMv2}_\textsc{base}$ can generate summaries containing correct clauses, but still have much inconsistency. 
The performance of both models indicates that Transformer is only capable of extracting important word-level information from dialogues in \textsc{DialogSum}, while $\textsc{UniLMv2}_\textsc{base}$ shows better performance on clause-level --- it can identify the speakers and partially preserve coreference information, consistent with findings of~\citet{levesque2012winograd} that pretraining is useful for coreference resolution.
However, it is far from human annotations.

\begin{table}[]
\centering

\setlength\tabcolsep{3pt}

\small
{
\begin{tabular}{l|cccc|ccc}
\hline
{\textbf{Model}}  & \multicolumn{4}{c|}{\textbf{Human Scores}} & \multicolumn{3}{c}{\textbf{ROUGE Scores}}  \\
& -1 & 0 & 1 & Avg.& R1 & R2 & RL\\
\hline
Transformer & 66\%&28\% & 6\% & -0.6 & 35.68 & 8.49 & 32.77 \\
\textsc{UniLMv2} &4\% & 56\% & 40\%& 0.36 & 47.46 & 21.33 & 44.93\\

\hline
\end{tabular}
}
\caption{Human evaluation on models' ability of preserving coreference information on \textsc{DialogSum}, with corresponding \textsc{ROUGE} scores.
Avg. stands for the averaged score here.
}
\label{coref}
\end{table}

\begin{table}[]
\centering
\setlength{\belowcaptionskip}{-0.3cm}

\small
{
\begin{tabular}{l|cc}
\hline
\textbf{Summary} & \multicolumn{2}{c}{\textbf{Human Scores}}   \\
& -1 &  1  \\
\hline
Summary 1 & 7.7\% & 92.3\% \\
Summary 2 & 20.5\%& 79.5\% \\
Summary 3 & 17.9\% & 82.1\% \\
Avg. & 15.4\% & 84.6\%\\
\hline
Transformer & 84.6\% & 15.4\% \\
\textsc{UniLMv2} & 30.8\% & 69.2\% \\

\hline
\end{tabular}
}
\caption{Human evaluation on models' ability of identifying interlocutors' intents.
}
\label{intents}
\end{table}

\begin{figure*}[!t]
    
    \includegraphics[width=1\textwidth]{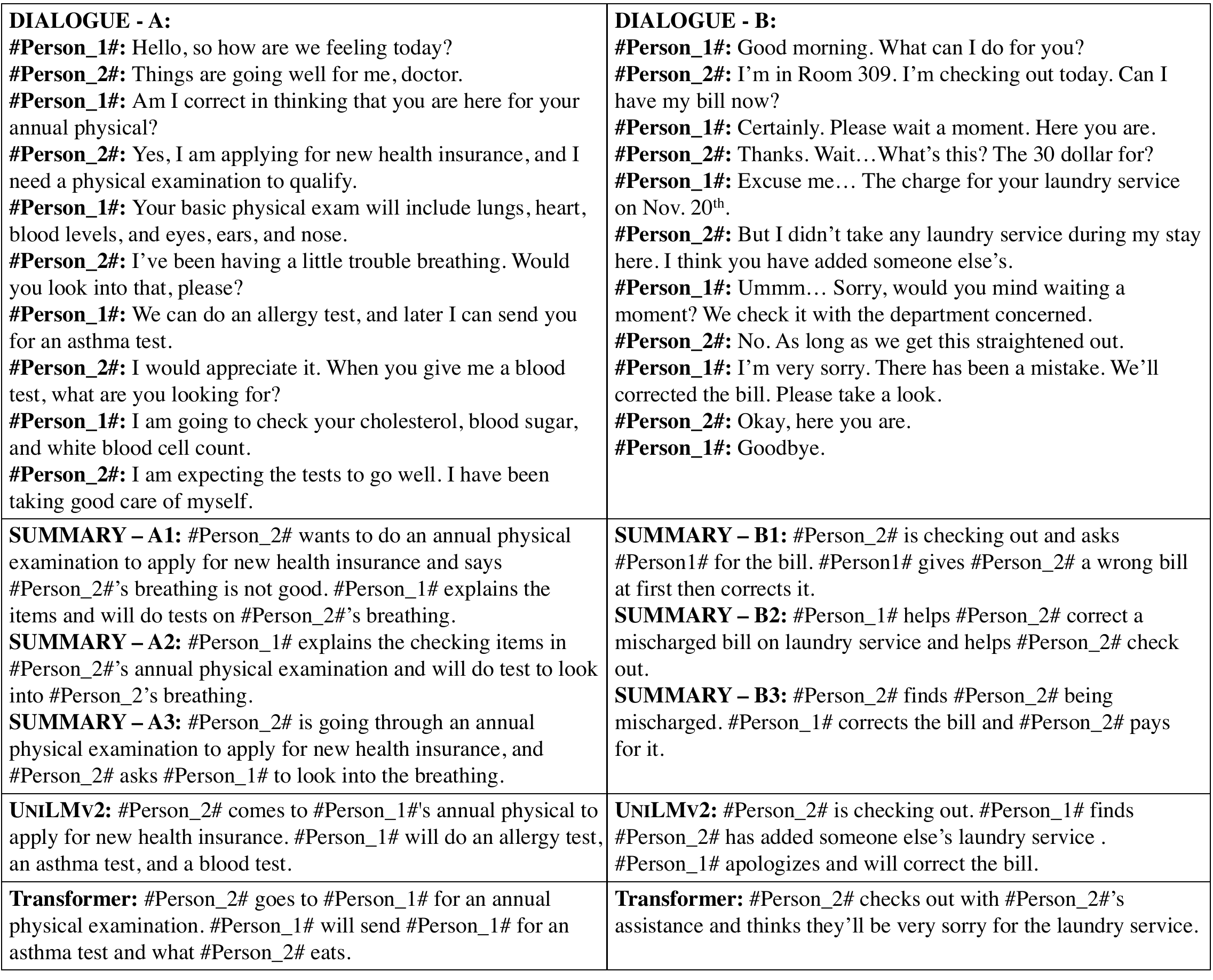}
    \caption{Case study on \textsc{DialogSum}. \textsc{Dialogue-A} - a doctor and a patient dialogue, \textsc{Dialogue-B} - a customer and a hotel service dialogue.}
    \label{casestudy}
\end{figure*}

\paragraph{Intent Identification}
As stated in Section~\ref{annotation}, we ask annotators to include important intents of interlocutors in their summaries, addition to the consequences of dialogues.
The intent here refers to the motivation of a speaker to initiate a conversation, e.g. ``\textit{want to do an annual physical}'' (c.f. Figure~\ref{casestudy}, \textsc{Dialogue-A}).
This can make summaries more comprehensive and readable.
Therefore, we conduct corresponding human evaluation on whether interlocutors' intents are described in summaries in the $50$ randomly selected data.

We first ask a judge to evaluate whether the intent is important to a dialogue, and we select $39$ dialogues that contain important intents.
Then, we ask the judge to give scores from $\{-1, 1\}$, where $1$ means that intents are identified correctly, $-1$ means incorrectly.
Note that we only focus on intent identification in the summary, and other errors should be ignored.
We also ask the judge to evaluate human annotated summaries.

The distribution of annotated scores is shown in Table~\ref{intents}. 
We see that most summaries generated by Transformer are scored as $-1$, which means that Transformer is incapable of generating summaries that correctly convey speakers' intents.
$\textsc{UniLMv2}_\textsc{base}$ shows much better performance, however, it is still below human performance.

\section{Challenges in \textsc{DialogSum}}
Compared to written texts, spoken dialogues can be more difficult for models to understand, and to summarize~\cite{goo2018abstractive}.
Therefore, we conduct error analysis and case studies on \textsc{DialogSum} to quantitatively and qualitatively discuss such challenges.
\subsection{Error Analysis}
We make error analysis on the $50$ selected model-generated summaries (Section~\ref{evaluationsection}).
Table~\ref{error} summarizes the five most frequent error types and their error rates.
In general, $\textsc{UniLMv2}_\textsc{base}$ shows better performance than Transformer, but its error rates are still high.
In particular, incorrect \textbf{coreference} (c.f. Section \ref{evaluationsection}) sees the highest error rates for both models, indicating that models can be confused because of interactive information flow.
Compared with Transformer, $\textsc{UniLMv2}_\textsc{base}$ can greatly avoid errors regarding unfactual information ($-52\%$) and syntactic ($-50\%$).
However, it still suffers from coreference issues, and tends to generate redundant summaries.

\begin{table}[]
\centering

\setlength\tabcolsep{3pt}

\small
{
\begin{tabular}{l|cc}
\hline
\textbf{Error Type} & \textbf{Transformer}  & \textbf{$\textsc{UniLMv2}_\textsc{base}$} \\
\hline
Incorrect Coref. & 94\% &  60\%\\
Missing Salient Inf. & 64\%  & 32\%\\
Redundant Inf. & 62\% & 44\% \\
Unfactual Inf. & 74\% & 22\% \\
Syntactic Error &72\% & 22\%\\
\hline
\end{tabular}
}
\caption{Error analysis of model performance on \textsc{DialogSum}. Coref. stands for coreference, and Inf. stands for Information.}
\label{error}
\end{table}


\subsection{Case Study}
We demonstrate two dialogues and their human-annotated/system-generated summaries in Figure~\ref{casestudy}.

First, a big challenge posed by spoken dialogues is that their \textbf{information flow} is different from monologic text, which is intuitively reflected in the dialogue \textbf{discourse structures}~\citep{wolf2005representing}. 
For example, two utterances can be closely related even where there is a large distance between them.
Such phenomena are common in spoken dialogues such as negotiations and procedures (e.g., medical consultation and police reports).
Due to the unique structure of the spoken dialogue, important information is rather dispersed than well-structured monologues and written-dialogues.


\textbf{Regular greetings} can be useless to written dialogue summaries (e.g. SAMSum), which is reflected by that \textsc{lead} is worse than \textsc{longest} on SAMSum (Table~\ref{bias}).
In contrast, \textsc{lead} outperforms \textsc{longest} by over $3\%$ on \textsc{DialogSum}. 
This is because, for spoken dialogues, such utterances sometimes express and indicate essential intents of speakers (c.f. Section~\ref{evaluationsection}).
\textbf{Farewells} also express the dialogue consequence and future plan of the speakers (e.g. dialogues in Figure~\ref{casestudy}).
Besides, \textbf{interruptions} appear frequently in the middle of conversations (Figure~\ref{casestudy}, \textsc{Dialogue-B}).
These interruptions make other speaker's utterances incomplete, adding redundant information, and can also destroy coherent discourse structures, making dialogues more difficult to encode.
These characteristics also make information in \textsc{DialogSum} dialogues more dispersed than existing datasets.

Second, \textbf{coreference} and \textbf{ellipsis} are frequent in spoken dialogues~\citep{grosz-etal-1995-centering, quan2019gecor}.
It is a natural behavior of communication that humans obey as a rhetorical principle for saving words and avoiding repetitions. 
Although it can be trivial for humans, their understanding can be challenging to a neural model.
For example, to correctly generate ``\textit{mischarged/wrong}'' in \textsc{summary-b1}-\textsc{summary-b3}, models need to understand ``\textit{I think you have added someone else's (laundry service on my bill)}'', where ``\textit{my bill}'' refers to ``\textit{\texttt{\#Person\_2\#}'s bill}''.

Third, \textbf{pragmatics} and \textbf{social common sense} give a unique challenge for spoken language understanding and has a significant impact on summarization.
From the last two sentences of \textsc{Dialogue-B}, human could understand that the ``\textit{Here you are}'' is actually ``\textit{make a payment}'', and ``\textit{Goodbye}'' indicates that the event ``\textit{check out}'' is finished.
It requires commonsense knowledge to fully understand such dialogues.
Beside, dialogues are summarized from a \textit{different} perspective (compared with speakers' perspective), which suggests that summarizing dialogues needs to go beyond summarizing dialogue contents, but also dialogue actions at the {\bf pragmatic} level.
For example, ``\textit{explains}'' in \textsc{summary-a1} and \textsc{summary-a2} summarizes multiple dialogue actions of \texttt{\#Person\_1\#}, ``\textit{agree}'' in Figure~\ref{example} (a) summarizes actions of both speakers.
It requires model to not only summarize \textit{what speakers are saying}, but also \textit{what they are doing}.


\section{Conclusion}
We presented \textsc{DialogSum}, a large-scale dialogue summarization dataset, investigating its characteristics and challenges empirically. 
Experiments with typical models show that \textsc{DialogSum} is highly abstractive, and poses unique challenges in discourse and complex co-references. 
From these observations, we made discussion on the uniqueness of spoken dialogue summarization, listing several key problems to consider in future modeling. 
To our knowledge, we are the first to release a large-scale dataset for real-life scenario dialogue summarization.

\section{Ethics Consideration}
As mentioned, we collect our data from DailyDialog, DREAM and MuTual that all are public for academic use.
The additional data are from www.tingroom.com, which are available to the public as well.
The sources of our dialogue data are freely accessible online without copyright constraint to academic use.


We hired annotators who have degrees in English Linguistics or Applied Linguistics.
Before formal annotation, we annotated 50 samples randomly extracted from the dataset, and calculated our average annotation time so we could set a fair salary for annotators' training annotation.
During the training annotation process, they were paid as well.
We also calculated the average annotation time for each dialogue during training, based on which we determined the final salary was around $9.5$ dollars per hour.
This hourly salary was the same for manual checking.
All of our annotators took this annotation as a part-time job.

\section*{Acknowledgments}
Yue Zhang is the corresponding author.
We would like to thank the anonymous reviewers for their comments, and Xuefeng Bai, Ming Shen and Bonnie Webber for insightful discussion and proofreading.
Also, we would like to thank Leyang Cui for sharing original dialogue data of MuTual dataset.
Our sincere appreciation goes to annotators who have contributed to building this dataset.
This work receives funding support from the Westlake University and Bright Dream Joint Institute for Intelligent Robotics, and a research grant from Rxhui Inc.

\bibliographystyle{acl_natbib}
\bibliography{anthology,acl2021}

\begin{thebibliography}{37}
\expandafter\ifx\csname natexlab\endcsname\relax\def\natexlab#1{#1}\fi

\bibitem[{Bao et~al.(2020)Bao, Dong, Wei, Wang, Yang, Liu, Wang, Gao, Piao,
  Zhou, and Hon}]{bao2020unilmv2}
Hangbo Bao, Li~Dong, Furu Wei, Wenhui Wang, Nan Yang, Xiaodong Liu, Yu~Wang,
  Jianfeng Gao, Songhao Piao, Ming Zhou, and Hsiao{-}Wuen Hon. 2020.
\newblock \href {http://proceedings.mlr.press/v119/bao20a.html} {Unilmv2:
  Pseudo-masked language models for unified language model pre-training}.
\newblock In \emph{Proceedings of the 37th International Conference on Machine
  Learning, {ICML} 2020, 13-18 July 2020, Virtual Event}, volume 119 of
  \emph{Proceedings of Machine Learning Research}, pages 642--652. {PMLR}.

\bibitem[{Bender and Koller(2020)}]{bender2020climbing}
Emily~M. Bender and Alexander Koller. 2020.
\newblock \href {https://doi.org/10.18653/v1/2020.acl-main.463} {Climbing
  towards {NLU}: {On} meaning, form, and understanding in the age of data}.
\newblock In \emph{Proceedings of the 58th Annual Meeting of the Association
  for Computational Linguistics}, pages 5185--5198, Online. Association for
  Computational Linguistics.

\bibitem[{Budzianowski et~al.(2018)Budzianowski, Wen, Tseng, Casanueva, Ultes,
  Ramadan, and Ga{\v{s}}i{\'c}}]{budzianowski2018multiwoz}
Pawe{\l} Budzianowski, Tsung-Hsien Wen, Bo-Hsiang Tseng, I{\~n}igo Casanueva,
  Stefan Ultes, Osman Ramadan, and Milica Ga{\v{s}}i{\'c}. 2018.
\newblock \href {https://doi.org/10.18653/v1/D18-1547} {{M}ulti{WOZ} - a
  large-scale multi-domain {W}izard-of-{O}z dataset for task-oriented dialogue
  modelling}.
\newblock In \emph{Proceedings of the 2018 Conference on Empirical Methods in
  Natural Language Processing}, pages 5016--5026, Brussels, Belgium.
  Association for Computational Linguistics.

\bibitem[{Carletta et~al.(2005)Carletta, Ashby, Bourban, Flynn, Guillemot,
  Hain, Kadlec, Karaiskos, Kraaij, Kronenthal et~al.}]{carletta2005ami}
Jean Carletta, Simone Ashby, Sebastien Bourban, Mike Flynn, Mael Guillemot,
  Thomas Hain, Jaroslav Kadlec, Vasilis Karaiskos, Wessel Kraaij, Melissa
  Kronenthal, et~al. 2005.
\newblock The ami meeting corpus: A pre-announcement.
\newblock In \emph{International workshop on machine learning for multimodal
  interaction}, pages 28--39. Springer.

\bibitem[{Chen and Yang(2020)}]{chen2020multi}
Jiaao Chen and Diyi Yang. 2020.
\newblock \href {https://doi.org/10.18653/v1/2020.emnlp-main.336} {Multi-view
  sequence-to-sequence models with conversational structure for abstractive
  dialogue summarization}.
\newblock In \emph{Proceedings of the 2020 Conference on Empirical Methods in
  Natural Language Processing (EMNLP)}, pages 4106--4118, Online. Association
  for Computational Linguistics.

\bibitem[{Cui et~al.(2020)Cui, Wu, Liu, Zhang, and Zhou}]{cui2020mutual}
Leyang Cui, Yu~Wu, Shujie Liu, Yue Zhang, and Ming Zhou. 2020.
\newblock \href {https://doi.org/10.18653/v1/2020.acl-main.130} {{M}u{T}ual: A
  dataset for multi-turn dialogue reasoning}.
\newblock In \emph{Proceedings of the 58th Annual Meeting of the Association
  for Computational Linguistics}, pages 1406--1416, Online. Association for
  Computational Linguistics.

\bibitem[{Dinan et~al.(2019)Dinan, Roller, Shuster, Fan, Auli, and
  Weston}]{dinan2018wizard}
Emily Dinan, Stephen Roller, Kurt Shuster, Angela Fan, Michael Auli, and Jason
  Weston. 2019.
\newblock \href {https://openreview.net/forum?id=r1l73iRqKm} {Wizard of
  wikipedia: Knowledge-powered conversational agents}.
\newblock In \emph{7th International Conference on Learning Representations,
  {ICLR} 2019, New Orleans, LA, USA, May 6-9, 2019}. OpenReview.net.

\bibitem[{Gehrmann et~al.(2018)Gehrmann, Deng, and Rush}]{gehrmann2018bottom}
Sebastian Gehrmann, Yuntian Deng, and Alexander Rush. 2018.
\newblock \href {https://doi.org/10.18653/v1/D18-1443} {Bottom-up abstractive
  summarization}.
\newblock In \emph{Proceedings of the 2018 Conference on Empirical Methods in
  Natural Language Processing}, pages 4098--4109, Brussels, Belgium.
  Association for Computational Linguistics.

\bibitem[{Gliwa et~al.(2019)Gliwa, Mochol, Biesek, and Wawer}]{gliwa2019samsum}
Bogdan Gliwa, Iwona Mochol, Maciej Biesek, and Aleksander Wawer. 2019.
\newblock \href {https://doi.org/10.18653/v1/D19-5409} {{SAMS}um corpus: A
  human-annotated dialogue dataset for abstractive summarization}.
\newblock In \emph{Proceedings of the 2nd Workshop on New Frontiers in
  Summarization}, pages 70--79, Hong Kong, China. Association for Computational
  Linguistics.

\bibitem[{Goo and Chen(2018)}]{goo2018abstractive}
Chih-Wen Goo and Yun-Nung Chen. 2018.
\newblock Abstractive dialogue summarization with sentence-gated modeling
  optimized by dialogue acts.
\newblock In \emph{2018 IEEE Spoken Language Technology Workshop (SLT)}, pages
  735--742. IEEE.

\bibitem[{Grosz et~al.(1995)Grosz, Joshi, and
  Weinstein}]{grosz-etal-1995-centering}
Barbara~J. Grosz, Aravind~K. Joshi, and Scott Weinstein. 1995.
\newblock \href {https://www.aclweb.org/anthology/J95-2003} {{C}entering: A
  framework for modeling the local coherence of discourse}.
\newblock \emph{Computational Linguistics}, 21(2):203--225.

\bibitem[{Hermann et~al.(2015)Hermann, Kocisk{\'{y}}, Grefenstette, Espeholt,
  Kay, Suleyman, and Blunsom}]{hermann2015teaching}
Karl~Moritz Hermann, Tom{\'{a}}s Kocisk{\'{y}}, Edward Grefenstette, Lasse
  Espeholt, Will Kay, Mustafa Suleyman, and Phil Blunsom. 2015.
\newblock \href
  {https://proceedings.neurips.cc/paper/2015/hash/afdec7005cc9f14302cd0474fd0f3c96-Abstract.html}
  {Teaching machines to read and comprehend}.
\newblock In \emph{Advances in Neural Information Processing Systems 28: Annual
  Conference on Neural Information Processing Systems 2015, December 7-12,
  2015, Montreal, Quebec, Canada}, pages 1693--1701.

\bibitem[{Kingma and Ba(2014)}]{kingma2014adam}
Diederik~P Kingma and Jimmy Ba. 2014.
\newblock Adam: A method for stochastic optimization.
\newblock \emph{arXiv preprint arXiv:1412.6980}.

\bibitem[{Koncel-Kedziorski et~al.(2019)Koncel-Kedziorski, Bekal, Luan, Lapata,
  and Hajishirzi}]{koncel2019text}
Rik Koncel-Kedziorski, Dhanush Bekal, Yi~Luan, Mirella Lapata, and Hannaneh
  Hajishirzi. 2019.
\newblock \href {https://doi.org/10.18653/v1/N19-1238} {{T}ext {G}eneration
  from {K}nowledge {G}raphs with {G}raph {T}ransformers}.
\newblock In \emph{Proceedings of the 2019 Conference of the North {A}merican
  Chapter of the Association for Computational Linguistics: Human Language
  Technologies, Volume 1 (Long and Short Papers)}, pages 2284--2293,
  Minneapolis, Minnesota. Association for Computational Linguistics.

\bibitem[{Kryscinski et~al.(2019)Kryscinski, Keskar, McCann, Xiong, and
  Socher}]{kryscinski2019neural}
Wojciech Kryscinski, Nitish~Shirish Keskar, Bryan McCann, Caiming Xiong, and
  Richard Socher. 2019.
\newblock \href {https://doi.org/10.18653/v1/D19-1051} {Neural text
  summarization: A critical evaluation}.
\newblock In \emph{Proceedings of the 2019 Conference on Empirical Methods in
  Natural Language Processing and the 9th International Joint Conference on
  Natural Language Processing (EMNLP-IJCNLP)}, pages 540--551, Hong Kong,
  China. Association for Computational Linguistics.

\bibitem[{Kryscinski et~al.(2020)Kryscinski, McCann, Xiong, and
  Socher}]{kryscinski-etal-2020-evaluating}
Wojciech Kryscinski, Bryan McCann, Caiming Xiong, and Richard Socher. 2020.
\newblock \href {https://doi.org/10.18653/v1/2020.emnlp-main.750} {Evaluating
  the factual consistency of abstractive text summarization}.
\newblock In \emph{Proceedings of the 2020 Conference on Empirical Methods in
  Natural Language Processing (EMNLP)}, pages 9332--9346, Online. Association
  for Computational Linguistics.

\bibitem[{Levesque et~al.(2012)Levesque, Davis, and
  Morgenstern}]{levesque2012winograd}
Hector Levesque, Ernest Davis, and Leora Morgenstern. 2012.
\newblock The winograd schema challenge.
\newblock In \emph{Thirteenth International Conference on the Principles of
  Knowledge Representation and Reasoning}. Citeseer.

\bibitem[{Lewis et~al.(2020)Lewis, Liu, Goyal, Ghazvininejad, Mohamed, Levy,
  Stoyanov, and Zettlemoyer}]{lewis2019bart}
Mike Lewis, Yinhan Liu, Naman Goyal, Marjan Ghazvininejad, Abdelrahman Mohamed,
  Omer Levy, Veselin Stoyanov, and Luke Zettlemoyer. 2020.
\newblock \href {https://doi.org/10.18653/v1/2020.acl-main.703} {{BART}:
  Denoising sequence-to-sequence pre-training for natural language generation,
  translation, and comprehension}.
\newblock In \emph{Proceedings of the 58th Annual Meeting of the Association
  for Computational Linguistics}, pages 7871--7880, Online. Association for
  Computational Linguistics.

\bibitem[{Li et~al.(2017)Li, Su, Shen, Li, Cao, and Niu}]{li2017dailydialog}
Yanran Li, Hui Su, Xiaoyu Shen, Wenjie Li, Ziqiang Cao, and Shuzi Niu. 2017.
\newblock \href {https://www.aclweb.org/anthology/I17-1099} {{D}aily{D}ialog: A
  manually labelled multi-turn dialogue dataset}.
\newblock In \emph{Proceedings of the Eighth International Joint Conference on
  Natural Language Processing (Volume 1: Long Papers)}, pages 986--995, Taipei,
  Taiwan. Asian Federation of Natural Language Processing.

\bibitem[{Likas et~al.(2003)Likas, Vlassis, and Verbeek}]{likas2003global}
Aristidis Likas, Nikos Vlassis, and Jakob~J Verbeek. 2003.
\newblock The global k-means clustering algorithm.
\newblock \emph{Pattern recognition}, 36(2):451--461.

\bibitem[{Lin(2004)}]{lin2004rouge}
Chin-Yew Lin. 2004.
\newblock \href {https://www.aclweb.org/anthology/W04-1013} {{ROUGE}: A package
  for automatic evaluation of summaries}.
\newblock In \emph{Text Summarization Branches Out}, pages 74--81, Barcelona,
  Spain. Association for Computational Linguistics.

\bibitem[{Liu(2019)}]{liu2019fine}
Yang Liu. 2019.
\newblock Fine-tune bert for extractive summarization.
\newblock \emph{arXiv preprint arXiv:1903.10318}.

\bibitem[{Liu and Lapata(2019)}]{liu2019text}
Yang Liu and Mirella Lapata. 2019.
\newblock \href {https://doi.org/10.18653/v1/D19-1387} {Text summarization with
  pretrained encoders}.
\newblock In \emph{Proceedings of the 2019 Conference on Empirical Methods in
  Natural Language Processing and the 9th International Joint Conference on
  Natural Language Processing (EMNLP-IJCNLP)}, pages 3730--3740, Hong Kong,
  China. Association for Computational Linguistics.

\bibitem[{Liu et~al.(2020)Liu, Shen, and Lapata}]{liu2020noisy}
Yang Liu, Sheng Shen, and Mirella Lapata. 2020.
\newblock Noisy self-knowledge distillation for text summarization.
\newblock \emph{arXiv preprint arXiv:2009.07032}.

\bibitem[{Miltsakaki et~al.(2004)Miltsakaki, Prasad, Joshi, and
  Webber}]{prasad2008penn}
Eleni Miltsakaki, Rashmi Prasad, Aravind Joshi, and Bonnie Webber. 2004.
\newblock \href {http://www.lrec-conf.org/proceedings/lrec2004/pdf/618.pdf}
  {The {P}enn {D}iscourse {T}reebank}.
\newblock In \emph{Proceedings of the Fourth International Conference on
  Language Resources and Evaluation ({LREC}{'}04)}, Lisbon, Portugal. European
  Language Resources Association (ELRA).

\bibitem[{Narayan et~al.(2018)Narayan, Cohen, and Lapata}]{narayan2018don}
Shashi Narayan, Shay~B. Cohen, and Mirella Lapata. 2018.
\newblock \href {https://doi.org/10.18653/v1/D18-1206} {Don{'}t give me the
  details, just the summary! topic-aware convolutional neural networks for
  extreme summarization}.
\newblock In \emph{Proceedings of the 2018 Conference on Empirical Methods in
  Natural Language Processing}, pages 1797--1807, Brussels, Belgium.
  Association for Computational Linguistics.

\bibitem[{Paulus et~al.(2018)Paulus, Xiong, and Socher}]{paulus2017deep}
Romain Paulus, Caiming Xiong, and Richard Socher. 2018.
\newblock \href {https://openreview.net/forum?id=HkAClQgA-} {A deep reinforced
  model for abstractive summarization}.
\newblock In \emph{6th International Conference on Learning Representations,
  {ICLR} 2018, Vancouver, BC, Canada, April 30 - May 3, 2018, Conference Track
  Proceedings}. OpenReview.net.

\bibitem[{Pennington et~al.(2014)Pennington, Socher, and
  Manning}]{pennington2014glove}
Jeffrey Pennington, Richard Socher, and Christopher~D Manning. 2014.
\newblock Glove: Global vectors for word representation.
\newblock In \emph{Proceedings of the 2014 conference on empirical methods in
  natural language processing (EMNLP)}, pages 1532--1543.

\bibitem[{Pilault et~al.(2020)Pilault, Li, Subramanian, and
  Pal}]{subramanian2019extractive}
Jonathan Pilault, Raymond Li, Sandeep Subramanian, and Chris Pal. 2020.
\newblock \href {https://doi.org/10.18653/v1/2020.emnlp-main.748} {On
  extractive and abstractive neural document summarization with transformer
  language models}.
\newblock In \emph{Proceedings of the 2020 Conference on Empirical Methods in
  Natural Language Processing (EMNLP)}, pages 9308--9319, Online. Association
  for Computational Linguistics.

\bibitem[{Quan et~al.(2019)Quan, Xiong, Webber, and Hu}]{quan2019gecor}
Jun Quan, Deyi Xiong, Bonnie Webber, and Changjian Hu. 2019.
\newblock \href {https://doi.org/10.18653/v1/D19-1462} {{GECOR}: An end-to-end
  generative ellipsis and co-reference resolution model for task-oriented
  dialogue}.
\newblock In \emph{Proceedings of the 2019 Conference on Empirical Methods in
  Natural Language Processing and the 9th International Joint Conference on
  Natural Language Processing (EMNLP-IJCNLP)}, pages 4547--4557, Hong Kong,
  China. Association for Computational Linguistics.

\bibitem[{Radev et~al.(2002)Radev, Hovy, and McKeown}]{radev2002introduction}
Dragomir~R. Radev, Eduard Hovy, and Kathleen McKeown. 2002.
\newblock \href {https://doi.org/10.1162/089120102762671927} {Introduction to
  the special issue on summarization}.
\newblock \emph{Computational Linguistics}, 28(4):399--408.

\bibitem[{Sandhaus(2008)}]{sandhaus2008new}
Evan Sandhaus. 2008.
\newblock The new york times annotated corpus.
\newblock \emph{Linguistic Data Consortium, Philadelphia}, 6(12):e26752.

\bibitem[{Sun et~al.(2019)Sun, Yu, Chen, Yu, Choi, and Cardie}]{sun2019dream}
Kai Sun, Dian Yu, Jianshu Chen, Dong Yu, Yejin Choi, and Claire Cardie. 2019.
\newblock \href {https://doi.org/10.1162/tacl_a_00264} {{DREAM}: A challenge
  data set and models for dialogue-based reading comprehension}.
\newblock \emph{Transactions of the Association for Computational Linguistics},
  7:217--231.

\bibitem[{Vaswani et~al.(2017)Vaswani, Shazeer, Parmar, Uszkoreit, Jones,
  Gomez, Kaiser, and Polosukhin}]{vaswani2017attention}
Ashish Vaswani, Noam Shazeer, Niki Parmar, Jakob Uszkoreit, Llion Jones,
  Aidan~N. Gomez, Lukasz Kaiser, and Illia Polosukhin. 2017.
\newblock \href
  {https://proceedings.neurips.cc/paper/2017/hash/3f5ee243547dee91fbd053c1c4a845aa-Abstract.html}
  {Attention is all you need}.
\newblock In \emph{Advances in Neural Information Processing Systems 30: Annual
  Conference on Neural Information Processing Systems 2017, December 4-9, 2017,
  Long Beach, CA, {USA}}, pages 5998--6008.

\bibitem[{Wolf and Gibson(2005)}]{wolf2005representing}
Florian Wolf and Edward Gibson. 2005.
\newblock \href {https://doi.org/10.1162/0891201054223977} {Representing
  discourse coherence: A corpus-based study}.
\newblock \emph{Computational Linguistics}, 31(2):249--287.

\bibitem[{Xu et~al.(2020)Xu, Gan, Cheng, and Liu}]{xu2020discourse}
Jiacheng Xu, Zhe Gan, Yu~Cheng, and Jingjing Liu. 2020.
\newblock Discourse-aware neural extractive text summarization.
\newblock In \emph{Proceedings of the 58th Annual Meeting of the Association
  for Computational Linguistics}, pages 5021--5031.

\bibitem[{Zhang et~al.(2018)Zhang, Dinan, Urbanek, Szlam, Kiela, and
  Weston}]{zhang2018personalizing}
Saizheng Zhang, Emily Dinan, Jack Urbanek, Arthur Szlam, Douwe Kiela, and Jason
  Weston. 2018.
\newblock \href {https://doi.org/10.18653/v1/P18-1205} {Personalizing dialogue
  agents: {I} have a dog, do you have pets too?}
\newblock In \emph{Proceedings of the 56th Annual Meeting of the Association
  for Computational Linguistics (Volume 1: Long Papers)}, pages 2204--2213,
  Melbourne, Australia. Association for Computational Linguistics.

\end{thebibliography}

\clearpage
\appendix

\section{Dialogue Topics}\label{a_topic}
\begin{figure*}[!t]

    \includegraphics[width=1\textwidth]{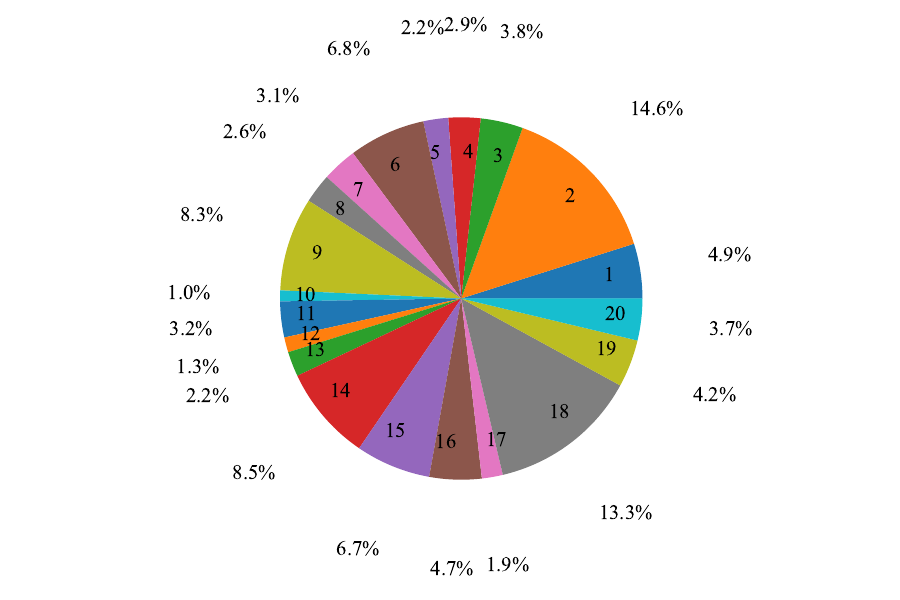}
    \caption{Proportion of dialogue topics.}
    \label{kmeans}
\end{figure*}

We use $k$-means~\citep{likas2003global} to cluster the dialogue topic with GloVe embedding~\citep{pennington2014glove}, where $k=20$.
Figure~\ref{kmeans} presents the proportion of clustering results.
Table~\ref{topicsss} presents the cluster topics with corresponding id, which is assigned by human.

\begin{table*}[]
\centering
\setlength\tabcolsep{3pt}
\small
{
\begin{tabular}{l|c|l|c|l|c|l|c}
\hline
\textbf{ID} & \textbf{topic} & \textbf{ID} & \textbf{topic} &\textbf{ID} & \textbf{topic} &\textbf{ID} & \textbf{topic} \\
\hline
1 & interpersonal relation & 6 & education & 11 & personal and business appoinment & 16 & transportation \\
2 & work and career & 7 & hobby & 12 & housing and apartment & 17 &in-store shopping \\
3 & causal chitchat& 8 & interview & 13 & consultation & 18& health and medicine\\
4 & hotel and restaurant service & 9 & vacation plan & 14 & personal life & 19 & entertainment\\
5 & sales & 10 & climate & 15 & economics & 20 & food ordering \\
\hline
\end{tabular}
}
\caption{Topic clusters of \textsc{DialogSum}.}
\label{topicsss}
\end{table*}

\section{Dialogue Turns}\label{turnnnnn}
The number of conversation turns can have a direct impact on neural models.
Multi-turn dialogues correspond to more complicated information flow and discourse structure. 
Following~\citet{chen2020multi}, we split test data based on dialogue turns, with a step size of $3$, and show model performance on different dialogue turns.

\begin{figure}[!t]
\setlength{\abovecaptionskip}{-0.1cm}
\setlength{\belowcaptionskip}{-0.3cm}

    \centering
    \includegraphics[width=1\columnwidth]{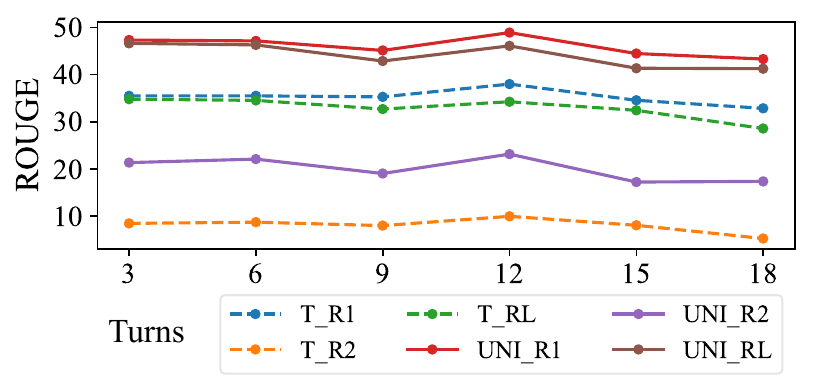}
    \caption{Model performance against the number of dialogue turns. 
    T - Transformer.
    UNI - \textsc{UniLMv2}.}
    \label{turns}
\end{figure}

The results are shown in Figure~\ref{turns}. 
The performance of Transformer and $\textsc{UniLMv2}_\textsc{base}$ decreases when number of turns grows, suggesting that more interactions between interlocutors and complicated discourse structures bring challenge. 
This phenomenon is also observed by~\citep{chen2020multi} for SAMSum.

\end{document}